\pdfoutput=1
\documentclass[11pt]{article}

\usepackage[preprint]{acl}

\usepackage{times}
\usepackage{latexsym}

\usepackage[T1]{fontenc}
\usepackage[utf8]{inputenc}

\usepackage{microtype}

\usepackage{inconsolata}
\usepackage[para]{threeparttable}
\usepackage{booktabs}
\usepackage{multirow}
\usepackage{graphicx}
\usepackage{amsmath}
\usepackage{amssymb}
\usepackage{xcolor}
\usepackage{hyperref}
\usepackage{mathtools}
\usepackage[ruled,vlined]{algorithm2e}
\usepackage{tcolorbox}
\usepackage{listings}
\usepackage{enumitem}
\usepackage{makecell}
\definecolor{c1}{HTML}{0049C0}
\usepackage{colortbl}
\usepackage{tabularx}
\definecolor{mygray}{gray}{.95}
\definecolor{mycell}{rgb}{0.85, 0.93, 0.97}
\definecolor{mycelltwo}{RGB}{255, 238, 241}
\usepackage{longtable}
\usepackage{arydshln}
\tcbuselibrary{breakable}
\usepackage{amssymb}
\makeatletter
\renewcommand{\maketag@@@}[1]{\hbox{\m@th\normalsize\normalfont#1}}%
\makeatother

\newcommand{\ie}{\textit{i.e.}}
\newcommand{\eg}{\textit{e.g.}}

\usepackage{graphicx}
\usepackage{bbding}
\usepackage{algpseudocode}

\DeclareMathOperator*{\argmax}{arg\,max}

\title{AdaRewriter: Unleashing the Power of Prompting-based Conversational Query Reformulation via Test-Time Adaptation}

\author{Yilong Lai,\hspace{1.5mm}
   Jialong Wu,\hspace{1.5mm}
   Zhenglin Wang,\hspace{1.5mm}
   Deyu Zhou\thanks{~~Corresponding Author.}\hspace{1.5mm}
   \\
        \hspace{0.5mm}School of Computer Science and Engineering, Key Laboratory of Computer Network\\
        and Information Integration, Ministry of Education, Southeast University, China
    \\
        \texttt{\{yilong.lai, jialongwu, zhenglin, d.zhou\}@seu.edu.cn} \\
}

\begin{document}
\maketitle
\begin{abstract}
Prompting-based conversational query reformulation has emerged as a powerful approach for conversational search, refining ambiguous user queries into standalone search queries. 
Best-of-N reformulation over the generated candidates via prompting shows impressive potential scaling capability.
% However, the training time tuning methods and the previous test time adaptation methods can not fully unleash their benefits.
However, both the previous tuning methods (training time) and adaptation approaches (test time) can not fully unleash their benefits.
In this paper, we propose AdaRewriter, a novel framework for query reformulation using an outcome-supervised reward model via test-time adaptation.
By training a lightweight reward model with contrastive ranking loss, AdaRewriter selects the most promising reformulation during inference.
Notably, it can operate effectively in black-box systems, including commercial LLM APIs.
Experiments on five conversational search datasets show that AdaRewriter significantly outperforms the existing methods across most settings, demonstrating the potential of test-time adaptation for conversational query reformulation.\footnote{The code are available in \url{https://github.com/init0xyz/AdaRewriter}}
\end{abstract}

\section{Introduction}
The rapid advancement of Large Language Models (LLMs) has driven significant innovations in information retrieval~\citep{zhao2023survey}. 
Notably, conversational AI search engines (\eg, \texttt{Perplexity}
% \footnote{\url{https://perplexity.ai/}} 
and \texttt{SearchGPT})
% \footnote{\url{https://searchgpt.com/}}) 
have attracted considerable attention due to their potential to shape the next generation of information retrieval~\citep{mo2024survey, mo_sigir_tutorial}.

A fundamental challenge of conversational search is understanding user intent by considering the historical context and the current query, as user inputs are often vague, ambiguous, or incomplete~\citep{conversational-search-book, mo2024survey}.
Two types of approaches have been proposed to tackle this challenge: 
(1) \underline{Conversation dense retrieval} involves training a dense encoder to generate conversational session embeddings~\citep{dense-retrieval-1, dense-retrieval-2, dense-retrieval-3, mo-etal-2025-uniconv, mao-etal-2024-chatretriever}. 
However, it can not be compatible with sparse retrieval systems like BM25 and may suffer from limited interpretability~\cite{cheng-etal-2024-interpreting}. 
(2) \underline{Conversational query reformulation} is explored to derive the user's search intent by turning the conversational context and current query into a standalone query.
With the advancement of LLMs, prompting-based query reformulation has emerged as a powerful way~\citep{LLM4CS, LLM-Aided, CHIQ}.
Previous studies have demonstrated the strong capability of the reformulation candidates generated through prompting, which have impressive potential scaling capability~\cite{CHIQ,AdaCQR}.
% These studies primarily investigate the fundamental capabilities of LLMs.
% , including instruction following, chain-of-thought, and self-consistency. 
% \jialong{add one sentence to introduce test-time scaling}
% Recent developments in test-time scaling, which enhances model performance by increasing computational resources during test-time, have further intensified interest in the community~\cite{scaling-works-1}.
% Motivated by the performance gains observed from aggregating multiple reformulation candidates~\cite{LLM4CS} and the potential in test-time scaling, w
% We aim to investigate the following research questions in this study: 
% \textbf{\textit{(i)}} Does scaling test-time computation significantly improve prompting-based query reformulation? 
% \textbf{\textit{(ii)}} What is the appropriate test-time scaling paradigm to unleash the power of prompting-based query reformulation?
% \textbf{\textit{(ii)}} Are there specific configurations that yield better performance when additional test-time computation is applied?
% \textbf{\textit{(ii)}} 
% \jialong{need revision to make problem be fundamental}
% However, based on our observation, we argue that the reasoning ability of LLMs in the context of conversational query reformulation remains insufficiently explored.

\begin{figure}[t]
    \centering
    \includegraphics[width=0.4\textwidth]{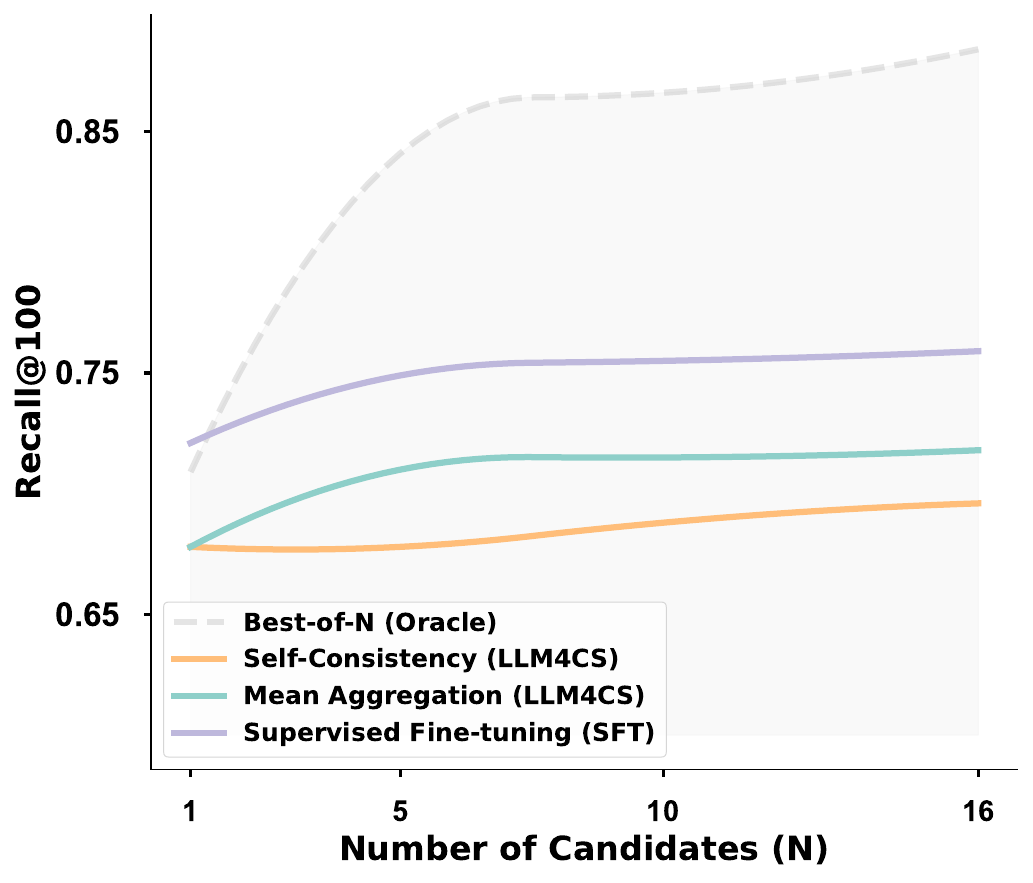}
    \vspace{-4mm}
    \caption{Comparison of training time and test-time adaptation strategies on the TopiOCQA using \texttt{LLaMA3.1-8B}.
    Best-of-N (Oracle) refers to prompting the model $N$ times and selecting the best-performing reformulation result.} 
    \label{fig:intro}
    \vspace{-4mm}
\end{figure}

As illustrated in Figure~\ref{fig:intro}, Best-of-N prompting-based reformulation demonstrates strong scalability.
However, simply supervised fine-tuning on the best reformulation at the training time has not yielded consistent performance gains, as described in Sec~\ref{appendix:comparision_with_sft}.
Another approach is to scale up during test time, leveraging increased computational resources to enhance model performance~\cite{scaling-works-1}.
~\citet{LLM4CS} investigate mean aggregation and self-consistency strategy~\cite{wang2023selfconsistency} during test time; they still exhibit a significant gap from the upper bound, as shown in Figure~\ref{fig:intro}.
This suggests the potential of test-time scaling has yet to be fully realized.
Based on these empirical observations, a natural question arises: \textit{How to design the appropriate \textbf{\underline{test-time scaling paradigm}} to unleash the power of prompting-based query reformulation?}

In this work, we introduce \textbf{AdaRewriter}, leveraging an outcome-supervised reward model to unleash the power of prompting-based conversational query reformulation. 
Inspired by the effectiveness of the reward model at test time ~\cite{google-prm-orm, MedAdapter}, a lightweight, BERT-sized reward model is proposed and trained using a contrastive ranking loss as the reward of reformulation in CQR is implicit.
During the inference stage, it serves as a scoring function to select the most promising reformulation.
It should be pointed out that AdaRewriter can be seamlessly applied in black-box conversational search systems, particularly those utilizing commercial LLMs via API services.

AdaRewriter achieves excellent performance on five widely used conversation search datasets, including TopiOCQA~\cite{topiocqa-datasets}, QReCC~\cite{qrecc-datasets}, and TREC CAsT 2019, 2020 \& 2021~\cite{cast19, cast20, cast21}.
Extensive experiments and analytical evaluations validate the effectiveness
and robustness of AdaRewriter.

The contributions of this paper are threefold:
\begin{itemize}[itemsep=0.6pt, parsep=0.1pt]
    \vspace{-4mm}
    \item To the best of our knowledge, we are the first
    to uncover and analyze the prompting-based query reformulation at test time under the Best-of-N paradigm.
    \item We propose AdaRewriter, a framework to unleash the power of prompting-based conversational query reformulation through an outcome-supervised reward model.
    \item Extensive experiments on several benchmark datasets demonstrate our proposed AdaRewriter outperforms existing methods across most settings, establishing its superiority in performance.
\end{itemize}

\section{Preliminaries}
\subsection{Task Formulation}
Conversational search systems aim to satisfy users' information-seeking needs in a multi-turn conversational form~\cite{conversational-search-book, mo2024survey}.
Formally, given the current query $q^k$ and historical context $\mathrm{H}^{k-1}=\{q^i,r^i\}_{i=1}^{k-1}$, the objective of these systems is to generate responses using the passages set $\mathrm{P}^k$ retrieved by an off-the-shelf retrieval system, where $k$ is the k-th turn of a conversation\footnote{For sake of convenience, we omit the superscript $k$ in the following sections.}. 

% However, user queries in conversational search systems are often complex, involving anaphora and ellipsis, which can obscure intent.
The conversational query reformulation task clarifies user intent by transforming the current query $q$ and historical context $\mathrm{H}$ into a standalone query $\mathcal{S}$.
% \jialong{make the two above sentences shorter}
Recent advancements in LLMs have made prompting-based CQR a promising approach, offering simplicity and superior performance.
In this method, the reformulated query $\hat{q}$ and the pseudo-response $\hat{r}$ are generated by LLM based on the task instructions $\mathcal{I}$ and few-shots examples $\mathcal{D}$, where each example consists of the whole conversation history and human-written turn-level query reformulation:
\begin{equation}
\{\hat{q}, \hat{r}\} = \operatorname{LLM}(\mathcal{I}, \mathcal{D}, \{q, \mathrm{H}\})
\end{equation}

\begin{figure*}[t]
    \centering
    \includegraphics[width=1.0\textwidth]{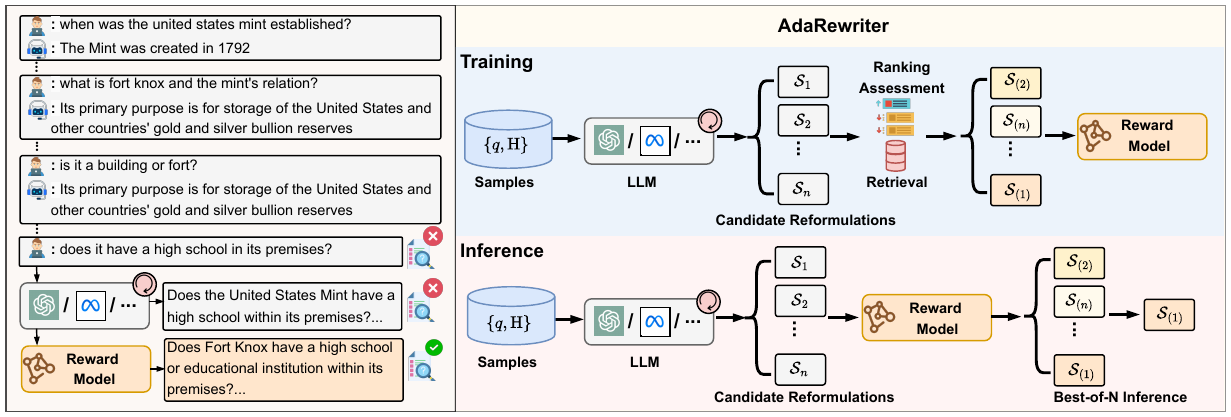}
    \caption{Overview of AdaRewriter.} 
    \label{fig:framework}
    \vspace{-4mm}
\end{figure*}

\subsection{Potential of Best-of-N in CQR}
\paragraph{Oracle}
We concatenate the reformulated query $\hat{q}$ with the pseudo-response $\hat{r}$ to form the reformulation query $\mathcal{S} = \hat{q}\oplus\hat{r}$, representing the user's search intent~\cite{ConvGQR}. 
To fully explore the potential of multiple candidates, we generate a set of reformulation queries $\{\mathcal{S}_1, \dots, \mathcal{S}_N\}$ and evaluate them using the Best-of-N paradigm, aiming to investigate the upper bound performance based on gold passage labels.  Figure~\ref{fig:intro} presents our preliminary results, indicating that the number of candidates improves performance. 

\paragraph{Training Time Fine-tuning}
Supervised fine-tuning(SFT) with the best-performing oracle reformulation via rejection sampling is a straightforward approach to further enhance the performance of prompting-based query reformulation. 
However, it does not consistently lead to performance gains based on our practices, as shown in Sec~\ref{appendix:comparision_with_sft}.

\paragraph{Test Time Adaptation}
Previous work~\cite{LLM4CS} proposes a simple yet effective method that generates multiple candidates query-response pairs $\{\hat{q}_1, \hat{r}_1\}, \{\hat{q}_2, \hat{r}_2\}, \dots, \{\hat{q}_N, \hat{r}_N\}$ and obtain the aggregated representation $\mathrm{s}$ in embedding space. 
Subsequently, the aggregated representation $\mathrm{s}$, treated as the standalone query $\mathcal{S}$, is utilized in dense retrieval systems to retrieve relevant passages. 
However, this method and self-consistency do not consistently lead to performance gains as the number of candidates increases, as shown in Figure~\ref{fig:intro}.

This motivates us to investigate prompting-based query reformulation further from the Best-of-N perspective.
Building on these insights and recent advancements in test-time scaling, we propose AdaRewriter, which leverages an outcome-supervised reward model to unleash the full potential of prompting-based query reformulation.

\section{Methodology}\label{sec:ORM}
% Following our observation,\jialong{need revise the motivation} 
To uncover the potential of prompting-based query reformulation under the Best-of-N paradigm, we propose AdaRewriter as presented in Figure~\ref{fig:framework}.
Specifically, we leveraged a vanilla LLM to generate reformulation candidates and construct implicit reward signals to train the reward model based on end-to-end performance assessment, as detailed in \S\ref{sec:training_paradigm}. 
\S\ref{sec:inference_paradigm} introduces the improved prompting-based query reformulation approach under the Best-of-N paradigm during inference.

\subsection{Reward Model Training}\label{sec:training_paradigm}
\paragraph{Constrative Ranking Loss}
Unlike traditional outcome-based methods that rely on binary classification labels, training a reward model for conversational query reformulation is non-trivial due to the absence of binary evaluation metrics in conversational search reformulation\footnote{We considered from end-to-end retrieval performance, as human-written labels are labor-intensive to collect and not always lead to the best performance.}. 
Without explicit reward, we leverage contrastive ranking loss, which is well-suited for tasks where relative ordering signals are much easier to obtain~\cite{liu-liu-2021-simcls, chuang-etal-2023-expand}. 
Specifically, the loss function targets to assign higher scores to top-ranked reformulations and lower scores to bottom-ranked ones:
\begin{equation}\label{eq:loss_function}
\mathcal{L} = \sum_{i=1}^n\sum_{j>i}\max(0,r_j - r_i +(j-i)\times\lambda)
\end{equation}
where $r_i$ is the score of candidate reformulation $\mathcal{S}_i$ with rank $i$ assigned by the trained reward model, $\lambda$ is a hyperparameter controls the margin between the candidates.
Despite the lack of explicit labels, this loss function can effectively optimize the model to distinguish the most promising reformulation $\mathcal{S}$ based on the assigned score among candidate reformulations.

\paragraph{Candidates Generation}
To construct candidate reformulations $\{ \mathcal{S}_1, \mathcal{S}_2, \cdots, \mathcal{S}_n \}$ described in Eq.~\eqref{eq:loss_function}, a vanilla LLM is employed, which generate multiple candidtates $\{S_{(1)}, S_{(2)}, \cdots, \mathcal{S}_{(n)}\}$ conditioned on a conversational session $\{ q, \mathrm{H}\}$.
The generation process is guided by instructions $\mathcal{I}$ and few-shot examples $\mathcal{D}$:
\begin{equation}\label{eq:candidates_generation}
\{S_{(1)}, S_{(2)}, \cdots, \mathcal{S}_{(n)}\} = \operatorname{LLM}(\mathcal{I}, \mathcal{D}, \{q,\mathrm{H}\})
\end{equation}

\paragraph{Ranking Assessment}
To rank the candidates, we utilize an end-to-end scoring function that combines multiple factors into a fusion score~\cite{reciprocal_rank_fusion, AdaCQR}:
\begin{equation}\label{eq:fusion_metric}
\operatorname{M}(\mathcal{S}_{(i)}) = \frac{1}{r_s(\mathcal{S}_{(i)}, p)} + \frac{1}{r_d(\mathcal{S}_{(i)}, p)}
\end{equation}
where $r_s(\mathcal{S}_{(i)}, p)$ denotes the corresponding rank with the gold passage $p$ giving query $\mathcal{S}_{(i)}$ in a dense retrieval system, and $r_s(\mathcal{S}_{(i)}, p)$ represents the rank in a sparse retrieval system. 
The candidate reformulation $\mathcal{S}_{(i)}$ is subsequently assigned a rank $j$ based on its performance according to the metric in Eq.~\eqref{eq:fusion_metric}, with higher ranks corresponding to better performance.

Therefore, the trained outcome-supervised reward model $g_\theta$ based on a BERT-sized model can be trained by the contrastive ranking Loss.
It can assess the quality of query $\mathcal{S}$ generated by LLM conditioned on a conversational session $\{q, \mathrm{H}\}$ and return a score $r$:
\begin{equation}
r = g_\theta(\mathcal{S}, \{q, \mathrm{H}\})
\end{equation}

\subsection{Best-of-N Inference}\label{sec:inference_paradigm}
Leveraging the outcome-supervised reward model $g_\theta$, our framework functions as a plug-and-play module to enhance prompting-based CQR during inference, adhering to the Best-of-N paradigm. 
Owing to test-time scalability, this module can be seamlessly integrated into a wide range of conversational search systems, regardless of whether the underlying large language model is deployed locally or accessed through commercial API services.

Specifically, given a conversational session $\{ q, \mathrm{H}\}$, the LLM generates multiple reformulation candidates $\{S_{(1)}, S_{(2)}, \cdots, \mathcal{S}_{(N)}\}$, as described in Eq.~\eqref{eq:candidates_generation}, where $N$ is the budget parameter that is adjustable during inference. 
The reward model $g_\theta$ then assigns scores to each candidate, and the highest-scoring candidate is selected as the most promising reformulation $\mathcal{S}$:
\begin{equation}
\mathcal{S} \leftarrow \mathcal{S}_{(k)}, k = \argmax_{j=1,\cdots,N}g_\theta(\mathcal{S}_{(j)}, \{q, \mathrm{H}\})
\end{equation}
The selected reformulation $\mathcal{S}$ is subsequently treated as the refined representation of the user's intent, leveraging the enhanced reasoning capabilities unlocked by our framework. The reformulation is then used to retrieve relevant passages, thereby improving the performance of conversational search systems.

\section{Experiments}
\paragraph{Datasets \& Evaluation Metrics} \quad The training data for the outcome-supervised reward model is derived from two widely used conversational search datasets: TopiOCQA~\cite{topiocqa-datasets} and QReCC~\cite{qrecc-datasets}. 
For evaluation, we use the test sets of TopiOCQA and QReCC. 
Additionally, to assess the zero-shot reformulation performance of our method, we conduct experiments on the TREC CAsT 2019, 2020, and 2021 datasets~\cite{cast19, cast20, cast21}. 
To evaluate the reformulation results, we adopt four standard metrics from information retrieval: MRR, NDCG@3, and Recall@10, which align with previous studies~\cite{cast20, ConvDR, ConvGQR}. 
Metric computation uses the \texttt{pytrec\_eval} tool~\cite{pytrec_eval}. 
Further details about the datasets can be found in the Appendix~\ref{sec:appendix-datasets}.

\paragraph{Implementation Details} In our prompting-based conversational query reformulation approach, we adopt the prompt used in ~\citet{LLM4CS}, specifically the "rewrite-and-response" setting with chain-of-thought, which represents the most advanced configuration. 
For the backbone selection in Sec~\ref{sec:training_paradigm}, we utilize \texttt{Llama2-7B} and \texttt{Llama3.1-8B} with a candidate size of $N = 16$ and a temperature setting of $0.7$, in line with previous studies~\cite{LLM4CS, CHIQ}. 
The outcome-supervised reward model is based on a lightweight BERT variant, \texttt{deberta-v3-base}.
% where the input consists of both the conversational session and the reformulation candidate. 
% The model outputs a score derived from the first CLS token's embedding. 
For retrieval, we employ BM25~\cite{bm25} for sparse retrieval and ANCE~\cite{ANCE} for dense retrieval, consistent with prior work~\cite{ConvGQR, LLM4CS}. 
The margin parameter $\lambda$ in Eq.~\eqref{eq:loss_function} is set to $0.1$, determined through grid search.
Further details about the implementation can be found in the Appendix~\ref{sec:appendix-implementation}.

\begin{table*}[t]
    \centering
    \small
    \begin{threeparttable}
    \begin{tabular*}{0.94\textwidth}{cllccccccc}
        \toprule
        & & &  \multicolumn{3}{c}{\textbf{TopiOCQA}} &  \multicolumn{3}{c}{\textbf{QReCC}}  \\ 
        \textbf{Type} & \textbf{Framework} & \textbf{Backbone} & \textbf{MRR} & \textbf{NDCG@3} & \textbf{R@10}  & \textbf{MRR} & \textbf{NDCG@3} & \textbf{R@10} \\
        \midrule
        \multirow{11}{*}{\rotatebox[origin=c]{90}{\textbf{Sparse (BM25)}}} 
        % & \texttt{Human Rewrite}  & - & - & - & - & 39.8 & 36.3 & 62.7 & 98.5 \\
        & T5QR  & T5-base & 11.3 & 9.8 & 22.1 & 33.4 & 30.2 & 53.8 \\
        & CONQRR & T5-base & - & - & - & 38.3 & - & 60.1 \\
        & EDIRCS & T5-base & - & - & - & 41.2 & - & 62.7 \\
        & ConvGQR & T5-base & 12.4 & 10.7 & 23.8 & 44.1 & 41.0 & 64.4 \\
        & IterCQR & T5-base & 16.5 & 14.9 & 29.3 & 46.7 & 44.1 & 64.4 \\
        & AdaCQR & T5-base & 17.8 & 15.8 & 34.1 & 52.4 & 49.9 & 70.9 \\ 
        & \textsc{RetPO} & Llama2-7B & \underline{28.3} & \underline{26.5} & 48.3 & 50.0 & 47.3 & 69.5\\
        & AdaCQR+Expansion & Llama2-7B$^*$ & \underline{28.3} & \underline{26.5} & \underline{48.9} & 55.1 & 52.5 & 76.5 \\
        \cmidrule(lr){2-10}
        & LLM-Aided & GPT3.5-Turbo & - & - & - & 49.4 & 46.5 & 67.1 \\
        & CHIQ-AD & Llama2-7B & 22.5 & 20.5 & 40.4 & 53.1 & 50.7 & 77.2 \\
        & CHIQ-Fusion & Llama2-7B$^*$ & 25.6 & 23.5 & 44.7 & 54.3 & 51.9 & \underline{78.5} \\
        % & LLama3.1-8B & & & & & & & \\
        & LLM4CS & Llama3.1-8B & 24.5 & 22.6 & 42.1 & 49.7 & 46.9 & 73.8 \\
        & \cellcolor{blue!5}{AdaRewriter (N=5)} & \cellcolor{blue!5}{Llama3.1-8B} & \cellcolor{blue!5}{28.2} & \cellcolor{blue!5}{26.2} & \cellcolor{blue!5}{48.3} & \cellcolor{blue!5}{54.0} & \cellcolor{blue!5}{51.3} & \cellcolor{blue!5}{77.4} \\
        & \cellcolor{blue!10}{AdaRewriter (N=16)} & \cellcolor{blue!10}{Llama2-7B} & \cellcolor{blue!10}{27.8} & \cellcolor{blue!10}{25.9} & \cellcolor{blue!10}{47.6} & \cellcolor{blue!10}{\underline{55.2}} & \cellcolor{blue!10}{\underline{52.8}} & \cellcolor{blue!10}{78.0} \\
        & \cellcolor{blue!20}{AdaRewriter (N=16)} & \cellcolor{blue!20}{Llama3.1-8B} & \cellcolor{blue!20}{\textbf{30.7}$^\dagger$} & \cellcolor{blue!20}{\textbf{28.8}$^\dagger$} & \cellcolor{blue!20}{\textbf{51.3}$^\dagger$} & \cellcolor{blue!20}{\textbf{56.2}$^\dagger$} & \cellcolor{blue!20}{\textbf{53.8}$^\dagger$} & \cellcolor{blue!20}{\textbf{78.8}$^\dagger$} \\
        % \cmidrule(lr){2-10}
        % & GPT4o-mini & & & & & & & \\
        % & \quad +LLM4CS & 29.2 & 27.8 & 48.2 & 72.3 & 50.3 & 47.7 & 75.0 & 92.8 \\
        % & \quad +AdaRewriter(N=5) & 31.1 & 29.6 & 51.4 & 74.3 & 53.8 & 51.5 & 77.8 & 94.0 \\
        \midrule 
        \multirow{14}{*}{\rotatebox[origin=c]{90}{\textbf{Dense (ANCE)}}}
        % & \texttt{Human Rewrite} & - & - & - & -  & 38.4 & 35.6 & 58.6 & 78.1\\
        % \cmidrule(lr){2-10} 
        & T5QR & T5-base & 23.0 & 22.2 & 37.6 & 34.5 & 31.8 & 53.1 \\
        & CONQRR & T5-base & - & - & - & 41.8 & - & 65.1 \\
        & EDIRCS & T5-base & - & - & - & 42.1 & - & 65.6 \\
        & IterCQR & T5-base & 26.3 & 25.1 & 42.6 & 42.9 & 40.2 & 65.5 \\
        & ConvGQR  & T5-base & 25.6 & 24.3 & 41.8 & 42.0 & 39.1 & 63.5 \\
        & AdaCQR & T5-base & 32.8 & 31.5 & 54.6 & 45.1 & 42.4 & 66.3 \\
        & \textsc{RetPO} & Llama2-7B & 30.0 & 28.9 & 49.6 & 44.0 & 41.1 & 66.7 \\
        & AdaCQR+Expansion & Llama2-7B$^*$ & 38.5 & 37.6 & 58.4 & 45.8 & 42.9 & 67.3 \\
        \cmidrule(lr){2-10} 
        & LLM-Aided & GPT3.5-Turbo & - & - & - & 43.5 & 41.3 & 65.6 \\
        & CHIQ-AD & Llama2-7B & 33.2 & 32.2 & 53.0 & 47.0 & 44.6 & 70.8 \\
        & CHIQ-Fusion & Llama2-7B$^*$ & 38.0 & 37.0 & \underline{61.6} & \underline{47.2} & 44.2 & \textbf{70.7} \\
        % & LLama3.1-8B & & & & & & & \\
        & LLM4CS(N=5) & Llama3.1-8B & 34.6 & 33.5 & 54.3 & 42.6 & 40.0 & 64.0 \\
        & LLM4CS(N=16) & Llama2-7B & 33.5 & 33.1 & 53.0 & 43.0 & 40.5 & 64.8 \\
        & LLM4CS(N=16) & Llama3.1-8B & 35.4 & 34.5 & 55.1 & 43.2 & 40.7 & 64.6 \\
        & \cellcolor{blue!5}{AdaRewriter (N=5)} & \cellcolor{blue!5}{Llama3.1-8B} & \cellcolor{blue!5}{\underline{38.9}} & \cellcolor{blue!5}{\underline{37.9}} & \cellcolor{blue!5}{59.6} & \cellcolor{blue!5}{46.1} & \cellcolor{blue!5}{43.4} & \cellcolor{blue!5}{69.2}\\
        & \cellcolor{blue!10}{AdaRewriter (N=16)} & \cellcolor{blue!10}{Llama2-7B} & \cellcolor{blue!10}{38.2} & \cellcolor{blue!10}{37.1} & \cellcolor{blue!10}{58.0} & \cellcolor{blue!10}{\underline{47.2}} & \cellcolor{blue!10}{\underline{44.4}} & \cellcolor{blue!10}{69.0}\\
        & \cellcolor{blue!20}{AdaRewriter (N=16)} & \cellcolor{blue!20}{Llama3.1-8B} & \cellcolor{blue!20}{\textbf{40.3}$^\dagger$} & \cellcolor{blue!20}{\textbf{39.7}$^\dagger$} & \cellcolor{blue!20}{\textbf{61.9}$^\dagger$} & \cellcolor{blue!20}{\textbf{47.5}} & \cellcolor{blue!20}{\textbf{44.7}$^\dagger$} & \cellcolor{blue!20}{\underline{69.8}} \\
        % \cmidrule(lr){2-10}
        % & GPT4o-mini & & & & & & & \\
        % & \quad +LLM4CS & 37.9 & 37.2 & 58.0 & 75.0 & 42.2 & 39.7 & 63.3 & 80.6 \\
        % & \quad +AdaRewriter(N=5) & 41.3 & 40.4 & 63.0 & 79.1 & 45.8 & 43.0 & 68.8 & 85.3 \\
        \bottomrule 
    \end{tabular*}
    \end{threeparttable}
    \vspace{-3mm}
    \caption{
    Evaluation results of various retrieval system types on the QReCC and TopiOCQA. 
    The best results among all methods are \textbf{bolded}, and the second-best results are \underline{underlined}.
    $^*$ denotes including fused results from a trained T5-based model.
    $\dagger$ denotes significant improvements with t-test at $p < 0.05$ over all compared baselines.
    }
    \vspace{-4mm}
    \label{table:main}
\end{table*}

\subsection{Baselines}
We conducted the primary experiments utilizing open-source large language models (LLMs) \texttt{Llama2-7B} and \texttt{Llama3.1-8B} to demonstrate the effectiveness of AdaRewriter.

Our approach is compared with various conversational query reformulation frameworks, which can be categorized into \textbf{fine-tuning} and \textbf{prompting-based} methods. 
The fine-tuning-based methods include T5QR~\cite{T5QR}, CONQRR~\cite{ConqRR}, EDIRCS~\cite{EDIRCS},  ConvGQR~\cite{ConvGQR}, IterCQR~\cite{IterCQR}, RetPO~\cite{RetPo}, and AdaCQR~\cite{AdaCQR}, while the prompting-based methods comprise LLM-Aided~\cite{LLM-Aided}, CHIQ~\cite{CHIQ}, and LLM4CS~\cite{LLM4CS}. 
Following ~\citet{CHIQ}, we also compare with the framework that fine-tuned LLM-based retrievers, including RepLLama~\cite{RepLLama}, E5-Mistral~\cite{e5-mistral}, and LLM-Embedder~\cite{LLM-Embedder}.
Additionally, we reproduce LLM4CS, a representative ensemble-based approach for CQR, which leverages the same LLM backbones as our method while varying the budget parameter $N$, to enable a fair and comprehensive comparison.

The Appendix~\ref{sec:appendix-baseline} presents comprehensive details of all the baseline methods. 
We also include the comparison with the Conversational Dense Retrieval(CDR) methods in Appendix~\ref{sec:appendix-comparsion-with-CDR}.

\begin{table*}[h]
    \centering
    \small
    \begin{threeparttable}
    \begin{tabular*}{2.01\columnwidth}{llccccccc}
        \toprule
        & & \multicolumn{2}{c}{\textbf{CAsT-19}} & \multicolumn{3}{c}{\textbf{CAsT-20}} & \multicolumn{2}{c}{\textbf{CAsT-21}} \\
        \textbf{Framework} & \textbf{Backbone} & \textbf{NDCG@3} & \textbf{R@10} & \textbf{MRR} & \textbf{NDCG@3} & \textbf{R@10} & \textbf{NDCG@3} & \textbf{R@10}\\
        \midrule
        T5QR & T5-base & 41.7 & - & 42.3 & 29.9 & - & 33.0 & - \\
        ConvGQR & T5-base & 43.4 & - & 46.5 & 33.1 & -  & 27.3 & - \\
        RepLLama & Llama2-7B & 31.6 & 10.6 &  26.8 & 18.3 & 10.4 & 32.7 & 19.6 \\
        E5-Mistral & Mistral2-7B & 31.3 & 9.5 & 22.0 & 15.4 & 8.4 & 32.5 & 20.5 \\
        LLM-Embedder & Llama2-7B & 36.6 & 11.4 & 25.2 & 15.4 & 8.7 & 31.2 & 17.3 \\
        AdaCQR+Expansion & Llama2-7B$^*$ & 48.5 & \textbf{13.0} & 56.6 & 38.5 & 19.2 & 45.6 & 25.0 \\
        \midrule
        % LLama3.1-8B &  &  &  &  &  &  &  &\\
        CHIQ-Fusion & Llama2-7B$^*$ & \textbf{50.5} & \underline{12.9} & 54.0 & 38.0 & 19.3  & 46.5 & 25.2\\
        LLM4CS (N=5) & Llama3.1-8B & 44.4 & 11.5 & 61.7 & 44.8 & 23.0 & \underline{50.5} & 25.7\\
        LLM4CS (N=10) & Llama3.1-8B & 45.5 & 11.9 & 61.9 & \underline{46.0} & 23.2 & \textbf{51.5} & 25.8\\
        \cellcolor{blue!5}AdaRewriter (N=5) & \cellcolor{blue!5}{Llama3.1-8B} & \cellcolor{blue!5}46.6 & \cellcolor{blue!5}{12.6} & \cellcolor{blue!5}{\underline{62.0}} & \cellcolor{blue!5}{45.6} & \cellcolor{blue!5}{22.6} & \cellcolor{blue!5}{49.5} & \cellcolor{blue!5}\underline{26.5}\\
        \cellcolor{blue!10}{AdaRewriter (N=10)} & \cellcolor{blue!10}{Llama2-7B} & \cellcolor{blue!10}{48.0} & \cellcolor{blue!10}{12.7} & \cellcolor{blue!10}{59.3} & \cellcolor{blue!10}{44.5} & \cellcolor{blue!10}{20.2} & \cellcolor{blue!10}{47.7} & \cellcolor{blue!10}{25.9} \\
        \cellcolor{blue!20}{AdaRewriter (N=10)} & \cellcolor{blue!20}{Llama3.1-8B} & \cellcolor{blue!20}{48.3} & \cellcolor{blue!20}{\textbf{13.0}} & \cellcolor{blue!20}{\textbf{63.0}$^\dagger$} & \cellcolor{blue!20}{\textbf{46.5}$^\dagger$} & \cellcolor{blue!20}{21.6} & \cellcolor{blue!20}{49.7} & \cellcolor{blue!20}\textbf{27.2}$^\dagger$\\
        \bottomrule 
    \end{tabular*}
    \end{threeparttable}
    \vspace{-2mm}
    \caption{
    Zero-shot experiment results on TREC CAsT 2019, 2020 \& 2021 datasets. The best results among all methods with similar settings are \textbf{bolded}, and the second-best results are \underline{underlined}.
    $^*$ denotes including fused results from a trained T5-based model.
    $\dagger$ denotes significant improvements with t-test at $p < 0.05$ over all compared baselines.
    }
    \vspace{-4mm}
    \label{table:zero-shot}
\end{table*}

\subsection{Main Results}
We evaluate our method on two benchmarks, TopiOCQA and QReCC, under both sparse and dense retrieval settings. 
As shown in Table~\ref{table:main}, AdaRewriter consistently outperforms baseline models across almost all scenarios.

On TopiOCQA with sparse retrieval, AdaRewriter (N=16) achieves MRR of 30.7, significantly outperforming LLM4CS's 24.5.
In the dense setting (ANCE), it also surpasses LLM4CS with an MRR of 40.3 vs. 35.4.
Performance further improves with larger candidate sets.
For example, on QReCC (sparse), MRR increases from 54.0 (N=5) to 56.2 (N=16). 
This suggests that AdaRewriter effectively utilizes candidate reformulations, thereby enhancing the model's ability to select the most promising one.
Similar trends are observed on the \texttt{Llama2-7B}.

Overall, AdaRewriter demonstrates strong adaptability to different retrieval conditions and benefits from scaling the number of candidate reformulations, offering an advantage in tasks requiring broader data exploration.

\subsection{Zero-shot Results}
In the zero-shot experiments conducted on the TREC CAsT 2019, 2020, and 2021 datasets, our proposed AdaRewriter consistently outperforms existing baselines across various budget parameters $N$, as shown in Table~\ref{table:zero-shot}.

Specifically, AdaRewriter achieves significant improvements on most metrics across all three datasets. 
For CAsT 2021, AdaRewriter yields strong gains in R@10, although its NDCG@3 performance is slightly lower. 
Despite this, our framework continues to exhibit considerable strength and robustness, confirming its capability to excel in retrieval performance and highlighting its robustness and adaptability across various datasets.
% Notably, AdaRewriter significantly improves key metrics such as MRR, NDCG@3, and R@10 across most datasets, further confirming its effectiveness and adaptability. 
% The results are presented in Table~\ref{table:zero-shot}.

% Based on the results, we observe that performance improvements are particularly noticeable when more candidates are used.
% Our framework also demonstrates strong compatibility with commercial LLMs, such as GPT4o-mini, highlighting its robust adaptability across different backbone models and budget settings.

% In detail, AdaRewriter exhibits impressive improvements in most metrics of the three CAsT datasets, consistently outperforming the baseline. 
% While its improvement on CAsT 2019 is solid, its performance over previous methods in terms of MRR and NDCG@3 is modest. 
% For CAsT 2021, AdaRewriter shows notable gains in R@10, while the performance in NDCG@3 is slightly lower. 
% Despite this, our framework continues to exhibit considerable strength and robustness, confirming its capability to excel in retrieval performance and highlighting its robustness and adaptability across various datasets.

% Overall, our framework delivers competitive results, demonstrating consistent advantages over baseline frameworks with notable improvements across most metrics. 
% These results further highlight the robustness and adaptability of our framework across various datasets.
\subsection{Comparison with Training-time Tuning}\label{appendix:comparision_with_sft}
To fully investigate the benefit of test-time adaptation, we compare our proposed AdaRewriter with three strong training-time baselines: supervised fine-tuning (SFT), SFT with Chain-of-Thoughts(CoT)~\cite{wei2022chain}, and direct preference optimization(DPO)~\cite{DPO}. 
All methods generate $N = 16$ candidate reformulations on the TopiOCQA dataset for a fair comparison.
SFT employs rejection sampling by selecting the best-performing candidates for fine-tuning. 
Building on vanilla SFT, we further incorporate chain-of-thought into the training labels, resulting in SFT with CoT.
DPO treats the best and worst candidates as chosen and rejected samples, respectively.

As shown in Table~\ref{table:comparison_with_sft}, AdaRewriter consistently outperforms the strong baselines in the datasets. 
Notably, on CAsT 2020, it achieves an MRR of 63.0, compared to 59.1 for SFT and 60.7 for DPO, demonstrating its robustness, especially on out-of-domain data.
These results highlight the effectiveness of test-time adaptation and confirm AdaRewriter's advantage in generating more relevant query reformulations.
We provide some details for the setup of SFT and DPO in the Appendix~\ref{appendix:training-time-details}.

\begin{table}[!tbp]
    \centering
    \small
    \resizebox{0.98\columnwidth}{!}{%
    \begin{tabular}{lccc}
    \toprule
    & \multicolumn{2}{c}{\textbf{TopiOCQA}} & \textbf{CAsT 20}  \\ 
    & MRR & R@10 & R@10 \\
    \midrule
    SFT & 39.2 & 59.4 & 59.1 \\
    SFT with CoT & 38.1 & 58.0 & 57.7 \\
    DPO & 39.1 & 59.8 & 60.7 \\
    AdaRewriter & 40.3 & 61.9 & 63.0\\
    \bottomrule
    \end{tabular}
    }
    \vspace{-2mm}
    \caption{Comparison with Training-time Tuning}
    \label{table:comparison_with_sft}
\end{table}

% These results underscore the superiority of test-time adaptation over training-time adaptation in the context of prompting-based query reformulation, confirming the effectiveness of AdaRewriter in finding more relevant query reformulations.

\section{Analysis}
In this section, we present a series of comprehensive experiments that aim to provide an in-depth analysis of the proposed AdaRewriter. 
Specifically, we investigate its effectiveness in addressing the following Research Questions (\textbf{RQs}):
\vspace{-2mm}

\begin{itemize}[itemsep=0.5pt, parsep=0.1pt]
\item \textbf{RQ1:} Can AdaRewriter be applied to black-box commercial LLMs?
\item \textbf{RQ2:} Does the conversational context $\mathrm{H}$ influence the score assigned to a reformulation query $\mathcal{S}$?
\item \textbf{RQ3:} How do the components (\eg, ranking loss, ranking assessment) impact the learning objectives of AdaRewriter?
\item  \textbf{RQ4:} Does AdaRewriter enhance the robustness of CQR in long conversations?
% \item  \textbf{RQ5:} What is the benefit of the test-time adaptation compared with training-time adaptations like SFT?
\end{itemize}
We also provide further discussions in Appendix~\ref{app:discussion}.

\subsection{Adaptation in Black-Box Models}
% \begin{table}[t]
%     \centering
%     \small
%     \begin{threeparttable}
%     \begin{tabular}{clccc}
%         \toprule
%         % & \multicolumn{3}{c}{\textbf{QReCC}}  \\ 
%         \textbf{Type} & \textbf{Frameworks} & \textbf{MRR} & \textbf{NDCG} &\textbf{R@10}\\
%         \midrule 
%         \multirow{2}{*}{\textbf{Sparse}}
%         % & Superior LLM Labels $Q^\star$  & 45.4 & 65.5\\
%         % \cmidrule(l){2-4}

%         &  LLM4CS & 29.2 & 27.8 & 48.2 \\
%         &  AdaRewriter & \textbf{31.1} & \textbf{29.6} & \textbf{51.4} \\

%         \midrule
%         \multirow{2}{*}{\textbf{Dense}}
%         & LLM4CS & 37.9 & 37.2 & 58.0 \\
%         & AdaRewriter & \textbf{41.3} & \textbf{40.4} & \textbf{63.0} \\
%         \midrule
%         \bottomrule 
%     \end{tabular}
%     \end{threeparttable}
%     \caption{
%     Evaluation results on TopiOCQA with budget parameter $N=5$, leveraging black-box model GPT4o-mini as backbone LLM. 
%     % \jialong{N=5}
%     }
%     \label{table:commerical-LLM}
% \end{table}

\begin{figure*}[t]
    \centering
    \includegraphics[width=0.8\textwidth]{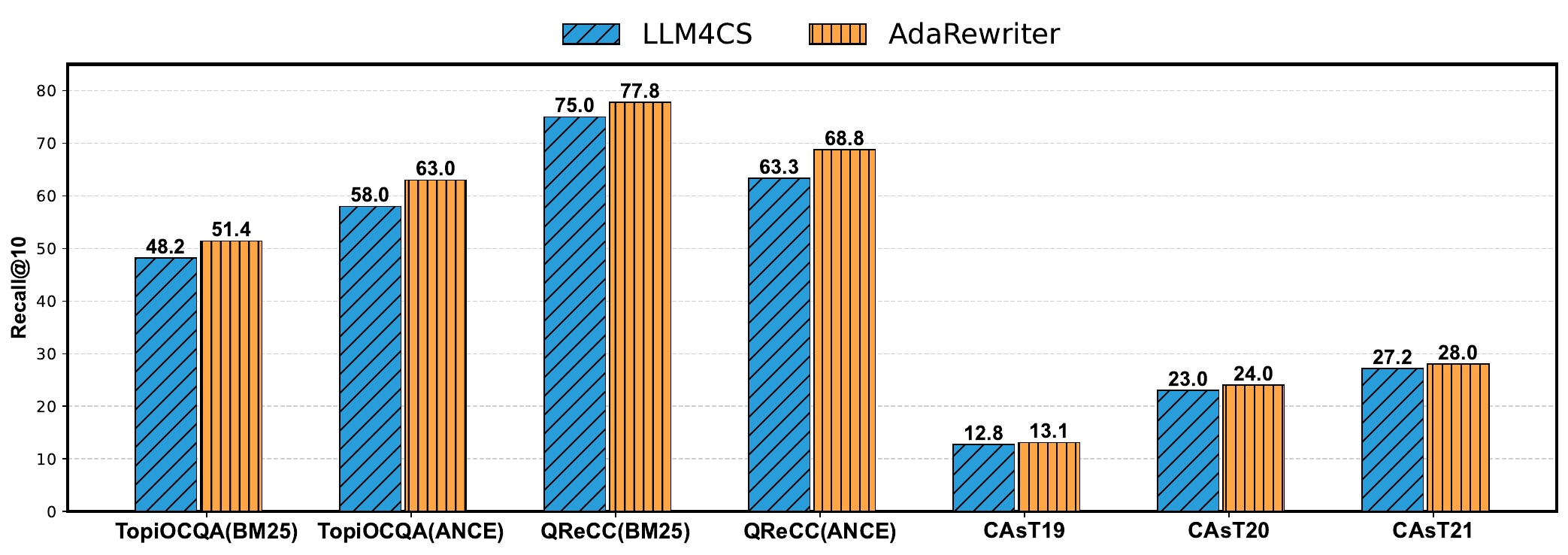}
    \vspace{-2mm}
    \caption{Performance comparsion on black-box model \texttt{GPT4o-mini}. We use $N=5$ for inference.} 
    \label{fig:4omini_comparison}
    \vspace{-4mm}
\end{figure*}

Building on the concept of test-time adaptation, our proposed AdaRewriter framework seamlessly integrates with conversational search systems that leverage commercial black-box LLMs, particularly those utilizing API services.

To answer \textbf{RQ1}, Figure~\ref{fig:4omini_comparison} presents evaluation results on the TopiOCQA, QReCC, and zero-shot datasets to validate AdaRewriter's effectiveness.
Experimental results show that AdaRewriter consistently enhances the performance of commercial LLMs, such as GPT4o-mini, across most evaluation metrics, even when trained on data generated by open-source LLMs. 
% Specifically, our framework significantly improves retrieval performance on TopiOCQA in both sparse and dense retrieval systems. 
For instance, compared to the baseline, AdaRewriter boosts the R@10 from 48.2 to 51.4 in sparse retrieval and from 58.0 to 63.0 in dense retrieval on the TopiCOQA dataset. 
Additionally, our framework demonstrates robust improvements on zero-shot datasets using commercial LLMs, as shown in Figure~\ref{fig:4omini_comparison}.

These results prove that AdaRewriter effectively boosts the commercial LLMs like GPT4o-mini, even with training data from open-source models, highlighting the robustness and promise of test-time adaptation for conversational query reformulation.
\vspace{-2mm}

\subsection{Contextual Dependency in Scoring}
\vspace{-2mm}
To investigate \textbf{RQ2}, we begin by examining the relationship between conversational history and reformulation query scoring. 
In conversational search systems, the meaning and relevance of a query can vary significantly depending on the context in which it is presented. 
Specifically, the conversational context $\mathrm{H}$ provides essential information about the ongoing conversation, such as user intent and topics, which may influence how a reformulated query is assessed.

To assess the impact of context $\mathrm{H}$ in our proposed framework, we conduct an ablation study in Table~\ref{table:ablation} (\colorbox{mycell}{w/o. Context $\mathrm{H}$}), in which the conversational context $\mathrm{H}$ is removed from the outcome-supervised reward model during both training and inference. 
The results reveal a significant drop in model performance when the context is excluded, showing the pivotal role of conversational context in guiding the outcome-supervised reward model's scoring of reformulated queries.
% Without the conversational context, the model struggles to capture the nuances of the conversation, leading to less accurate scoring of reformulation queries. 
% This performance degradation underscores the pivotal role of conversational context in guiding the outcome-supervised reward model's scoring of reformulated queries.

\begin{figure*}[t]
    \centering
    \includegraphics[width=0.8\textwidth]{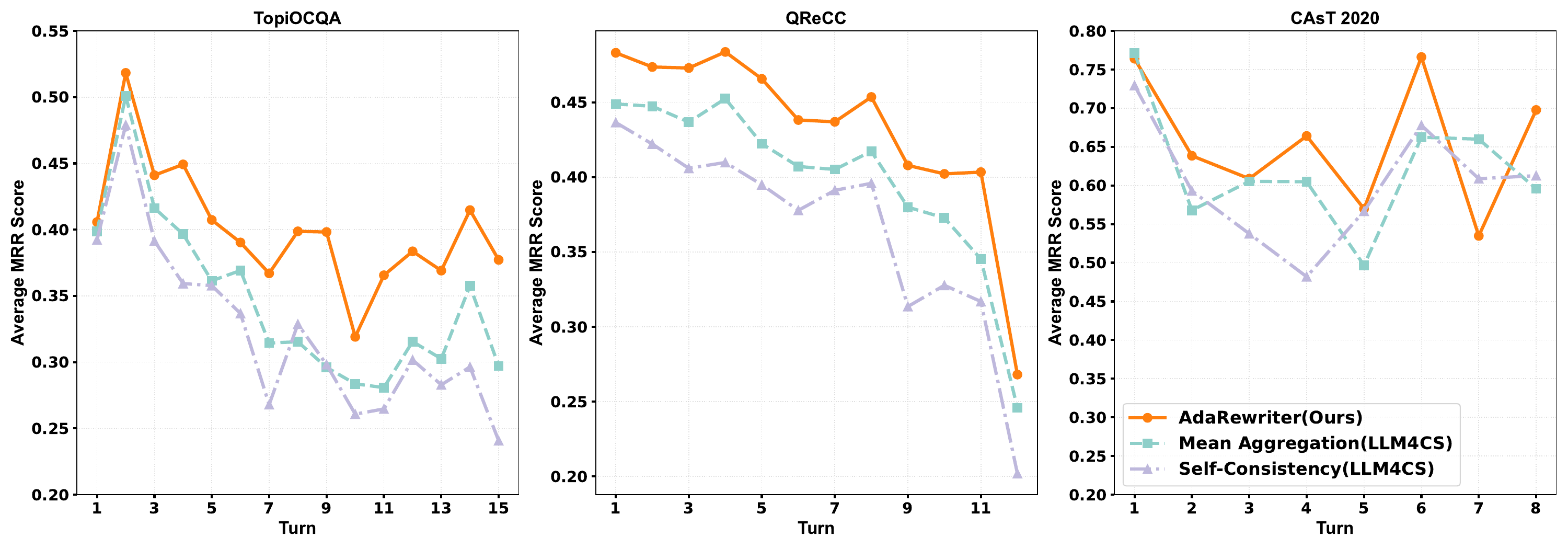}
    \vspace{-4mm}
    \caption{Turn-round performance comparison on TopiOCQA, QReCC, and TREC CAsT 2020.} 
    \label{fig:topic_shift}
    \vspace{-5mm}
\end{figure*}

\subsection{Influence of the Learning Objective}
\begin{table}[t]
    \centering
    \small
    \begin{threeparttable}
    \begin{tabular*}{0.98\columnwidth}{clccc}
        \toprule
        % & \multicolumn{3}{c}{\textbf{QReCC}}  \\ 
        \textbf{Type} & \textbf{Abaltion Variants} & \textbf{MRR} & \textbf{R@10}\\
        \midrule 
        \multirow{4}{*}{\rotatebox[origin=c]{90}{\textbf{Sparse}}}
        & AdaRewriter (\textit{Ours}) & \textbf{30.7} & \textbf{51.3}\\
        \cmidrule(r){2-4}
        & \cellcolor{mycell}{\quad  w/o. Context $\mathrm{H}$} & \cellcolor{mycell}{27.3} & \cellcolor{mycell}{44.9} \\
        & \cellcolor{mycelltwo}{\quad w/o. Ranking Loss} & \cellcolor{mycelltwo}{24.6} & \cellcolor{mycelltwo}{43.0} \\
        & \cellcolor{mygray}{\quad w/o. Rank Assessment} & \cellcolor{mygray}{23.8} & \cellcolor{mygray}{41.8} \\
        % \cdashline{2-4}[1pt/4pt]
        % & \quad w/o. Labels $Q^\star$ & 44.9 & 63.7 \\
        \midrule
        \multirow{5}{*}{\rotatebox[origin=c]{90}{\textbf{Dense}}} 
        & AdaRewriter (\textit{Ours}) & \textbf{40.3} & \textbf{61.9}\\
        \cmidrule(r){2-4}
        & \cellcolor{mycell}{\quad w/o. Context $\mathrm{H}$} & \cellcolor{mycell}{36.2} & \cellcolor{mycell}{56.4} \\
        & \cellcolor{mycelltwo}{\quad w/o. Ranking Loss} & \cellcolor{mycelltwo}{34.4} &  \cellcolor{mycelltwo}{53.2}\\
        & \cellcolor{mygray}{\quad w/o. Ranking Assessment} & \cellcolor{mygray}{32.8} & \cellcolor{mygray}{51.5} \\
        \midrule
        \bottomrule 
    \end{tabular*}
    \end{threeparttable}
    \caption{
    Ablation study for the learning objective and contextual dependency of AdaRewriter on TopiOCQA dataset. We use \texttt{LLama3.1-8B} and $N=16$ for inference.
    }
    \vspace{-4mm}
    \label{table:ablation}
\end{table}

% To better understand the contributions of each component to our reward model’s learning objectives, as highlighted by \textbf{RQ3}, we conduct an ablation study to assess the effectiveness of each individual component.
To investigate the individual contributions of our reward model's learning objectives as addressed in \textbf{RQ3}, we conduct an ablation study.

Specifically, we evaluate two variants: (1) \colorbox{mycelltwo}{w/o Ranking Loss}, where the ranking loss is replaced by a cross-entropy loss assigning the true label the top rank and the false label to the bottom; and (2) \colorbox{mygray}{w/o Ranking Assessment}, where candidate reformulations are randomly ordered instead of ranked.
% We design two variants to isolate the impact of different factors: (1) \colorbox{mycelltwo}{w/o. Ranking Loss}: This variant replaces the ranking loss with cross-entropy loss, where the true label is assigned the top rank and the false label the bottom rank. (2) \colorbox{mygray}{w/o. Ranking Assessment}: In this variant, the ranking of candidate reformulations is replaced by a random order.

Table~\ref{table:ablation} shows the results of these variants. 
Notably, the MRR in the dense retrieval drops from 40.3 to 34.4 when the ranking loss is removed, and also decreases to 32.8 when the ranking assessment is omitted. These findings demonstrate that both the contrastive loss and the ranking assessment are crucial for achieving strong performance, highlighting the importance of our proposed learning objectives for the reward model.

\subsection{Robustness in Long Conversation}
One of the primary challenges in conversational search systems is sustaining performance in extended conversation, as highlighted by 
\textbf{RQ4}. 
To answer this question, we assess the robustness of our proposed method across three datasets, which include TopiOCQA, QReCC, and TREC CAsT 2020. 
The results, presented in Figure~\ref{fig:topic_shift}, reveal that as the length of the conversation increases, performance across all methods experiences a notable decline. This suggests that long conversations still present a challenge for current CQR methods.

Despite this general decline in performance, AdaRewriter consistently outperforms the other baselines across all conversation turns. 
Notably, even as the dialogue length increases, AdaRewriter maintains a higher performance compared to Mean Aggregation and Self-Consistency proposed by~\citet{LLM4CS}, which demonstrates a more substantial drop in effectiveness. 
This behavior suggests that AdaRewriter is more robust to the degradation typically observed in long conversations.
\vspace{-2mm}
% Our findings underscore that AdaRewriter contributes positively to the robustness of conversational search systems, particularly in long conversation scenarios. 
% This characteristic positions it as a promising solution for real-world conversational systems, where maintaining query relevance and accuracy over multiple turns is crucial.

% \vspace{-3.5mm}
\section{Related Works}
\paragraph{Conversational Query Reformulation}
Query reformulation plays a crucial role in conversational search systems, addressing the inherent complexity of user intent, which often involves semantic challenges such as anaphora and ellipsis~\cite{conversational-search-book,mo2024survey}. 
Current conversational query reformulation adopts hybrid approaches that combine query rewriting and query expansion, as exemplified by~\citet{ConvGQR}.
In the era of LLMs, prompting-based query reformulation has garnered significant attention due to its simplicity and superior performance. 
\citet{LLM-Aided} treats LLMs as both query rewriters and rewrite editors, following a ``\textit{rewrite-then-edit}'' paradigm to refine reformulations. 
\citet{LLM4CS} further explores advanced prompting strategies, such as few-shot learning, chain-of-thought reasoning, and self-consistency, demonstrating the remarkable efficacy of prompting-based approaches.
~\citet{ivica2024} leverages the beam search score of multiple rewrites and aggregates them with their scores for both sparse and dense retrieval in an unsupervised manner.
Building on these developments, \citet{CHIQ} proposed a two-step method that leverages the basic capabilities of open-source LLMs to enhance the conversational history for conducting query reformulation.

% Inspired by recent advancements in test-time adaptation and scaling, we propose a test-time adapter to optimize prompting-based query reformulation further, unlocking the reasoning potential of LLMs in the context of conversational search.

\paragraph{Test-time Supervision and Scaling}
Enhancing LLMs through test-time supervision and scaling test-time computation represents a promising direction for building robust and self-improving agent systems~\cite{scaling-works-1}. 
A series of works have focused on improving the reasoning capabilities of LLMs by incorporating reward model supervision during test-time inference~\cite{google-prm-orm, openai-verify-step-by-step}. 
In addition to these methods, test-time supervision has been proposed to improve the performance of LLMs in specific target domains using lightweight adapters~\cite{bbox-adapter, hydra, MedAdapter}. 
For example,~\citet{MedAdapter} employs a lightweight model to rank outputs generated by LLMs in the medical domain, enhancing the domain-specific performance.

However, based on our empirical observations, the ability of LLMs in the context of conversational search remains insufficiently explored.
% Apart from this, it's non-trivial to use test-time supervision and scaling the test-time computation as the nature of gold query reformulation does not exist for conversational search systems. 
To address this limitation, we propose leveraging a contrastive ranking loss to effectively train a lightweight reward model, unlocking LLM's reasoning capability in conversational search.
To the best of our knowledge, we are the first to uncover and analyze the prompting-based conversational query reformulation at test time under the Best-of-N paradigm.

\section{Conclusion}
In this paper, we aim to unleash the power of prompting-based query reformulation at test time within the Best-of-N paradigm. 
Therefore, we propose AdaRewriter, a framework that effectively uses a lightweight outcome-supervised reward model as a scoring function to select the most promising reformulation. 
Extensive experimental evaluations across several benchmark datasets demonstrate that AdaRewriter consistently outperforms existing methods in most settings. 
These contributions advance the understanding of user intent in conversational search systems and improve the effectiveness of prompting-based query reformulation.

\section*{Limitation}
% In this work, we demonstrate the effectiveness of prompting-based conversational query reformulation using an outcome-supervised reward model under the Best-of-N paradigm. 
We identify the below limitations in AdaRewriter:

% While the reward model is lightweight and the latency of AdaRewriter is comparable to the previous work~\cite{LLM4CS}, the primary bottleneck in latency arises from generating multiple reformulation candidates through the LLMs. 
% Nevertheless, we believe that enhancing prompting-based query reformulation via test-time adaptation holds promise due to its simplicity and effectiveness. This could reduce the requirement for extensive passage re-ranking.
% Additionally, test-time adaptation and scaling are promising, as the Best-of-N paradigm has shown superior performance across various tasks. 
% Our approach's latency can also be reduced by applying existing inference acceleration works~\cite{Fast_BoN, wang-etal-2025-seed}.

% Furthermore, there is a crucial trade-off between computational cost (i.e., increasing the number of candidates, $N$) and latency. A more effective and robust strategy may involve allocating more computational resources to harder scenarios and fewer resources to easier ones, which will be the focus of our future work.

% Lastly, due to budget constraints, while we have demonstrated the effectiveness of our method on black-box commercial LLMs, we have been unable to report the performance of AdaRewriter with a larger candidate number.

Although the reward model is lightweight and the latency of AdaRewriter is comparable to that of previous work~\cite{LLM4CS}, the primary latency bottleneck stems from the process of generating multiple reformulation candidates using LLMs. 
Despite this, we believe that improving prompting-based query reformulation through test-time adaptation shows considerable potential, as it combines both simplicity and effectiveness. 
This approach may reduce the need for extensive passage re-ranking. 
Additionally, test-time adaptation and scaling offer promising results, particularly with the Best-of-N paradigm, which has demonstrated superior performance across various tasks~\cite{scaling-works-1}.

To further reduce latency, our method could benefit from applying existing inference acceleration techniques~\cite{Fast_BoN, wang-etal-2025-seed}. 
A key trade-off also exists between computational cost and latency, specifically when increasing the number of candidates $N$. 
A more efficient strategy may involve dynamically allocating computational resources based on reformulation task difficulty, \ie, generating more candidates for complex scenarios and fewer for simpler ones. 

Lastly, due to budget constraints, while we have demonstrated the effectiveness of AdaRewriter on black-box commercial LLMs, we have been unable to evaluate its performance with a larger candidate set $N$.
% \newpage

\section*{Acknowledgments}
The authors would like to thank the anonymous reviewers for their insightful comments. This work is funded by the National Natural Science Foundation of China (Grant No.62176053). 
This work is supported by the Big Data Computing Center of Southeast University.

\bibliography{custom}
\clearpage
\newpage
\appendix

\begin{table*}[t]
    \centering
    \small
    \begin{threeparttable}
    \begin{tabular}{lcccccc}
        \toprule
        \textbf{Framework} & \textbf{TopiOCQA} & \textbf{QReCC} &
      \textbf{CAsT-19} & \textbf{CAsT-20} & \textbf{CAsT-21} & \textbf{Avg.}\\
        \midrule
        Conv-ANCE~\cite{ANCE} & 20.5 & 45.6 & 34.1 & 27.5 & 34.2 & 32.4 \\
        ConvDR~\cite{ConvDR} & 26.4 & 35.7 & 43.9 & 32.4 & 37.4 & 35.2 \\
        Conv-SPLADE~\cite{splade} & 29.5 & 46.6 & - & 28.1 & 29.9 & - \\
        InstructoR-ANCE~\cite{instructor} & 23.7 & 40.5 & -  & 29.6 & 34.9 & - \\
        LeCoRE~\cite{LeCoRe} & 32.0 & 51.1 & 42.2 & 37.7 & 50.8 & 42.8 \\
        ConvAug~\cite{ConvAug} & 33.3 & 50.4 & - & 30.7 & 36.8 & -  \\
        ChatRetriever~\cite{mao-etal-2024-chatretriever} & 40.1 & 52.5 & 52.1 & 40.0 & 49.6 & 46.9 \\
        \rowcolor{blue!10}AdaRewriter (LLama3.1-8B, N=5) & 37.9 & 51.3 & 46.6 & 45.6 & 49.5 & 46.2\\
        \rowcolor{blue!20}AdaRewriter (LLama3.1-8B, N=16) & 39.7 & 53.8 & 48.3 & 46.5 & 49.7 & \underline{47.6}\\
        \rowcolor{blue!30}AdaRewriter (GPT4o-mini, N=5) & 40.4 & 51.5 & 49.0 & 47.3 & 52.5 & \textbf{48.1}\\
        \midrule
        \bottomrule 
    \end{tabular}
    \end{threeparttable}
    \caption{
    NDCG@3 performance comparison of our proposed AdaRewriter and Conversational Dense Retrieval(CDR) methods.
    % As several methods do not have results for the CAsT-19 dataset, we do not report MRR performance on CAsT-19, but our method achieves the \textbf{highest} MRR score of 74.5.
    The best average results among all methods are \textbf{bolded}, and the second-best results are \underline{underlined}.
    }
    \label{table:cdr_comparsion}
\end{table*}

\section{Discussion}
\label{app:discussion}
% \jialong{add discussion of evaluation on rag or cqr in appendix}
% \subsection{AdaRewriter as Reformulation Evaluator}
% The evaluation of query reformulation primarily relies on two main approaches~\cite{mo2024survey}:
% (1) \textbf{lexical overlapping}, which assesses the accuracy of the reformulated query relative to the reference query by computing token-level precision, recall, and F1 score, and 
% (2) \textbf{end-to-end evaluation}, which measures the effectiveness of the reformulated query based on its final retrieval performance.
% However, these evaluation methods have their limitations: while lexical overlapping is efficient, it provides only an indirect measure and does not reflect the real effectiveness of the reformulated queries in downstream tasks; end-to-end assessment, though comprehensive, is computationally intensive and may be influenced by model biases.

% In contrast, our proposed neural-based reward model could be a trade-off evaluation suite between efficiency and accuracy, demonstrating its robustness across both sparse and dense retrieval systems.
% Based on our practices, the reward model could serve as an effective proxy for assessing the quality of query reformulation, with potential applications in retrieval-augmented generation (RAG) systems~\cite{wu2025webwalker}, conversational search systems, and hard negative mining for retriever training.

\subsection{Comparsion with AdaCQR}
AdaCQR~\cite{AdaCQR} aims to improve the performance of conversational query reformulation through a two-stage training paradigm. In the first stage, the model is trained using a large set of pseudo-labels generated by a large language model. 
The second stage further refines the model via iterative self-training with a contrastive ranking loss.

Despite demonstrating effectiveness, AdaCQR faces two notable limitations:
\begin{itemize}
    \item AdaCQR exhibits a performance gap compared to LLM-based methods. To enable a fair comparison with such methods, an additional query expansion step using an LLM is required (i.e., the AdaCQR+Expansion setting proposed in the original paper).
    \item AdaCQR functions primarily as a training-time alignment approach, which restricts its applicability in real-world scenarios, particularly in environments where LLMs are accessed as black-box systems.
\end{itemize}

To address these limitations, AdaRewriter is proposed as a lightweight framework that employs a reward model to select the most promising candidate reformulations by combining query rewriting and expansion. 
It retains simplicity while benefiting from the concept of test-time scaling.

Moreover, AdaRewriter demonstrates the potential of leveraging test-time scaling and test-time adaptation in the context of conversational query reformulation. 
We believe this could offer some insights for future research in the field of conversational search.

\subsection{Comparsion with CDR Methods}\label{sec:appendix-comparsion-with-CDR}

Conversational Dense Retrieval(CDR) represents an orthogonal approach to conversational query reformulation in the context of conversational search. This methodology focuses on training dense retrievers to improve the representation of both the current query and its associated historical context. Although a direct comparison may not be appropriate, we present a performance comparison between our proposed AdaRewriter and several CDR methods evaluated across the QReCC, TopiOCQA, and TREC CAsT datasets, as shown in Table~\ref{table:cdr_comparsion}.

We compare AdaRewriter with the following representative CDR methods: Conv-ANCE~\cite{ANCE}, ConvDR~\cite{ConvDR}, Conv-SPLADE~\cite{splade}, InstructorR-ANCE~\cite{instructor}, LeCoRE~\cite{LeCoRe}, ConvAug~\cite{ConvAug}, and ChatRetriever~\cite{mao-etal-2024-chatretriever}. 
Among these, ChatRetriever stands out as one of the most representative works in the era of LLMs, which fine-tunes an LLM using contrastive learning and leverages the conversational session's embeddings to retrieve relevant passages. 
The results in Table~\ref{table:cdr_comparsion} demonstrate that our proposed method achieves consistently strong performance across all five datasets, highlighting the robustness and effectiveness of AdaRewriter.

Moreover, conversational query reformulation-based approaches, such as AdaRewriter, offer superior explainability compared to CDR methods. 
This is valuable for enhancing user intent understanding and shows promise for improving conversational search systems.

\section{Experimental Details}\label{sec:appendix-experimental}
\subsection{Datasets Details}\label{sec:appendix-datasets}
\begin{table}[t]
    \centering
    \small
    \resizebox{\columnwidth}{!}{%
    \begin{tabular}{lcccc}
    \toprule
    & \multicolumn{2}{c}{\textbf{QReCC}} &  \multicolumn{2}{c}{\textbf{TopiOCQA}}  \\ 
    & Train & Test & Train & Test \\
    \midrule
    \#\ Dialogues & 10823 & 2775 & 3509 & 205 \\
    \#\ Turns & 29596 & 8209 & 45450 & 2514\\
    \midrule
    \#\ Collections & \multicolumn{2}{c}{54M} & \multicolumn{2}{c}{25M} \\
    \midrule
    \bottomrule
    \end{tabular}
    }
    \caption{The statistics of QReCC and TopiOCQA datasets. }
    \label{table:qrecc_topiocqa_details}
\end{table}

\begin{table}[h]
    \centering
    \small
    \resizebox{\columnwidth}{!}{%
    \begin{tabular}{lccc}
    \toprule
    & \textbf{CAsT-19} & \textbf{CAsT-20} & \textbf{CAsT-21}  \\ 
    \midrule
    \# Dialogues & 50 & 25 & 26 \\
    \# Turns & 479 & 208 & 239 \\
    \midrule
    \# Collections & 38M & 38M & 42M \\
    \midrule
    \bottomrule
    \end{tabular}
    }
    \caption{The statistics of TREC CAsT 2019, 2020, and 2021 datasets. }
    \label{table:cast_dataset_details}
\end{table}

This paper uses five datasets: TopiOCQA~\cite{topiocqa-datasets}, QReCC~\cite{qrecc-datasets}, and TREC CAsT 2019~\cite{cast19}, 2020~\cite{cast20}, and 2021~\cite{cast21}. 
TopiOCQA and QReCC contain both training and testing data, while TREC CAsT datasets provide only testing data for zero-shot experiments.

The QReCC dataset consists of 14K conversations with 80K question-answer pairs, and we aim to retrieve relevant passages from a collection of 54M passages. 
The TopiOCQA dataset contains 3.9K topic-switching conversations with 51K question-answer pairs, with a passage collection of 25M passages. 
Detailed statistics for both datasets are shown in Table~\ref{table:qrecc_topiocqa_details}.

TREC CAsT 2019, 2020, and 2021 are known for their complexity in conversational search under a zero-shot setting. 
Table~\ref{table:cast_dataset_details} provides more details.

\subsection{Implementation Details}\label{sec:appendix-implementation}
All experiments are conducted on a server with four Nvidia GeForce 3090 GPUs. 

Our framework is implemented using the Huggingface Transformers\footnote{\url{https://github.com/huggingface/transformers}} and PyTorch Lightning\footnote{\url{https://github.com/Lightning-AI/pytorch-lightning}}. 
The AdamW optimizer is used with a learning rate of 5e-6, following a cosine learning rate schedule with a warmup ratio of 0.1. 
Training is carried out for 10 epochs, and model checkpoints are saved at the end of each epoch. 
We employed the vLLM~\cite{VLLM} framework for candidate construction and inference, ensuring reproducibility by saving the results for inference.
The retrieval systems were implemented using Faiss~\cite{faiss} and Pyserini~\cite{pyserini}. For the BM25 algorithm, we set the parameters as follows: $k_1 = 0.82, b = 0.68$ in QReCC, and $k_1 = 0.9, b = 0.4$ in TopiOCQA. Here, $k_1$ controls non-linear term frequency normalization, while $b$ adjusts the scaling of the inverse document frequency. The query length was set to 32, and the concatenated reformulation query length was set to 256, following prior works~\cite{LLM4CS}.

\subsection{Training-time Tuning Details}\label{appendix:training-time-details}
We use \texttt{Llama-Factory}~\cite{llamafactory} to conduct experiments on supervised fine-tuning (SFT) and direct preference optimization (DPO). 
To accommodate our hardware constraints, we adopt the LoRA technique with the rank $r=16$. 
The training is performed for 3 epochs with a learning rate of 1e-4.

\section{Baseline Details}\label{sec:appendix-baseline}
We compare AdaRewriter with the following representative baselines in the CQR task:
\begin{itemize}[itemsep=0.5pt, parsep=0pt, left=-0.1em]
\item \textbf{T5QR}~\cite{T5QR} is a vanilla baseline that train the T5-base~\cite{T5-paper} model to perform CQR tasks.

\item \textbf{CONQRR}~\cite{ConqRR} aligns the T5-base reformulation model with retrievers through direct optimization using reinforcement learning.

\item \textbf{ConvGQR} ~\cite{ConvGQR} improves retrieval performance by utilizing two fine-tuned T5-base models, with one dedicated to query reformulation and the other to query expansion.

\item \textbf{EDIRCS} ~\cite{EDIRCS} effectively generates reformulation queries by combining non-autoregressive text-selection techniques with autoregressive token generation, utilizing a fine-tuned T5-base model.

\item \textbf{LLM-Aided} ~\cite{LLM-Aided} employs ChatGPT ~\cite{openaigpt35} to conduct query reformulation via a ``rewrite-then-edit'' prompting strategy.

\item \textbf{IterCQR} ~\cite{IterCQR} aligns the T5-base reformulation model with the dense retriever by minimizing Bayesian risk, which is driven by the semantic similarity between the query and the gold passage.

\item \textbf{\textsc{RetPO}}~\cite{RetPo} leverages large language models to produce diverse reformulations through multi-perspective prompting, generates binarized comparisons informed by retriever feedback, and fine-tunes LLama2-7B via direct preference optimization~\cite{DPO}.

\item  \textbf{\textsc{AdaCQR}}~\cite{AdaCQR} aligns the reformulation model with sparse and dense retrieval systems through a fusion metric, demonstrating the effectiveness of guiding reformulation using hybrid retrieval signals. The reformulation model leverages the T5-base and uses a vanilla LLama2-7B for query expansion.

\item  \textbf{CHIQ}~\cite{CHIQ} utilizes the fundamental capabilities of LLMs to improve the quality of contextual history, thereby enhancing query rewriting performance. For comparison, we employ the most advanced CHIQ-Fusion, which combines reformulated queries generated by a fine-tuned T5-based model and the LLama2-7B model, utilizing result-level fusion techniques to derive the final retrieval outcomes.

\item \textbf{LLM4CS}~\cite{LLM4CS} is our primary comparison method, exploring various strategies for aggregating multiple reformulation candidates to improve conversational search. The candidates are generated by a large language model through direct prompting. We employ the most advanced settings, including Rewrite-and-Response (RAR), Mean Aggregation, and Chain-of-Thought (CoT). To ensure a fair comparison, we present results based on our reproduction of the officially released code.
\end{itemize}

\begin{table*}
\begin{tcolorbox}[pad at break*=1mm,colback=white!95!gray,colframe=gray!50!black,title=Prompt for Conversational Query Reformulation]
\textbf{Instruction}
\begin{lstlisting}[breaklines=true, xleftmargin=0pt, breakindent=0pt, columns=fullflexible]
For an information-seeking dialog, please help reformulate the question into rewrite that can fully express the user's information needs without the need of context, but also generate an informative response to answer the question. I will give you several example multi-turn dialogs, where each turn contains a question as well as a rewrite and a response that you need to generate. The rewrite part begins with a sentence explaining the reason for the generated rewrite.
\end{lstlisting}

\textbf{Demonstrations}
\begin{lstlisting}[breaklines=true, xleftmargin=0pt, breakindent=0pt, columns=fullflexible]
Example #1:
Question: What should I consider when buying a phone?
Rewrite: This is the first turn. So the question should be rewritten as: What should I consider when buying a phone?
Response: The design of the phone and the overall look and feel of the phone are very important. You should be comfortable with the way the phone looks and feels when you hold it in your hand. In addition, don't be afraid to get a phone from a different manufacturer than you're used to. Consider an older, used, or refurbished phone to save money. Sometimes a year-old or even a two-year-old device might give you everything you need.

Question: Cool. Which one would you recommend?
Rewrite: Based on Turn 1, you are inquiring about what should be considered when buying a phone. So the question should be rewritten as: Cool. Which smartphone would you recommend for me?
Response: Just because a phone has everything and the kitchen sink doesn't mean you have to buy it - it's just an option for those who might have a larger budget. The same logic applies to pretty much anything you can buy. If you want something more affordable, cheaper, but has the latest in Android and that comes with 5000mAh of battery, then the Motorola Moto G7 Power is an excellent phone to consider. If you want an Android smartphone that's cheaper and more affordable, and you still find the Moto G7 Power a bit off your budget, then I highly recommend the Samsung Galaxy A10e.

(*@\textcolor{gray}{<--- Omit Following Demonstrations --->}@*)
\end{lstlisting}

\textbf{Annotated Sample}
\begin{lstlisting}[breaklines=true, xleftmargin=0pt, breakindent=0pt, columns=fullflexible]
Context:(*@\textcolor{c1}{
\{\{current\_context\}\}
}@*)

Current Question: (*@ \textcolor{c1}{
\{\{current\_query\}\}
}@*)

Now, you should give me the rewrite and response of the **Current Question** under the **Context**. The output format should always be: "Rewrite: $Reason. So the question should be rewritten as: $Rewrite\nResponse: $Response." Note that you should always try to rewrite it and generate an informative response. Never ask for clarification or say you don't understand it in the generated rewrite and response. Go ahead!
\end{lstlisting}
\end{tcolorbox}
\caption{The prompt used to obtain rewritten query $\hat{q}$ and pseudo-response $\hat{r}$.}
\label{prompt_for_qrecc}
\end{table*}

\section{Case Study}\label{sec:appendix-case}
\begin{table*}[t!]
    \centering
    \resizebox{\textwidth}{!}{
    \def\arraystretch{0.9}
    \begin{tabularx}{\textwidth}{X}
        \toprule
            \textbf{Conversation}:\\
            Q1: what is roosevelt dime? \\
            A1: Is the current dime, or ten-cent piece, of the United States \\
            Q2: who designed it? \\
            A2: John R. Sinnock \\
            Q3: who is he? \\
            A3: Eighth Chief Engraver of the United States Mint \\
            Q4: mention a few controversies of his? \\
            A4: The letters "JS" actually stood not for John Sinnock, but for Joseph Stalin. The urban folk story coincided with the Second Red Scare. Another controversy was an allegation that Sinnock copied or borrowed the design of the President's profile from a bronze bas relief created by sculptress Selma H. Burke \\
            Q5: did the aforementioned mint struck the roosevelt dime? \\
            A5: UNANSWERABLE \\
            Q6: who made the dime? \\
            A6: John R. Sinnock \\
            Q7: when was the united states mint established? \\
            A7: The Mint was created in 1792 \\
            Q8: where is its location?\\
            A8: In Philadelphia \\
            Q9: what is fort knox and the mint's relation? \\
            A9: Its primary purpose is for storage of the United States and other countries' gold and silver bullion reserves. \\
            Q10: is it a building or fort? \\
            A10: Is a fortified vault building \\
            Q11: any recent incident happened over there related to shooting? \\
            A11: On 18 October 1993, Arthur Hill went on a shooting rampage, killing three and wounding two before attempting suicide, shooting and severely wounding himself. \\
            Q12: how does air corps utilize it? \\
            A12: As a training base during World War II. \\
            Q13: is it used for protecting valuable objects? \\
            A13: For protection after the Japanese attack on Pearl Harbor in 1941, the Declaration of Independence, the Constitution of the United States and the Gettysburg Address were all moved for safekeeping
            \textbf{Original Query}: does it have a high school in its premises? (\textbf{rank: Not Found}) \\\\
            \textbf{Max-prob Rewritten Query}: Does the United States Mint have a high school within its premises? The United States Mint does not have...(\textbf{rank: Not Found}) \\\\
            \textbf{AdaRewriter}(\textit{Ours}): Does \underline{Fort Knox} have a high school or educational institution within its premises? Fort Knox does not have a high school ...(\textbf{rank: 1})\\\\
            \textbf{Gold Passage}: \underline{Fort Knox} is one of only four Army posts (the others being Fort Campbell, Kentucky, Fort Meade, Maryland, and Fort Sam Houston, Texas) that still has a high school located on-post. Fort Knox High School was built in 1958 and has undergone only a handful of renovations...\\\\
        \bottomrule 
    \end{tabularx}}
    \caption{
    Successful case study on TopiOCQA (id: 126\_14). The \underline{underline} part shows the decontextualized information in the reformulation query.
    }
    \label{table:case_study_topiocqa_success}
\end{table*}

To show the effectiveness of AdaRewriter, we provide a detailed case in Table~\ref{table:case_study_topiocqa_success}.

\end{document}